\documentclass[letterpaper, 10 pt, conference]{ieeeconf}  

\IEEEoverridecommandlockouts                              

\usepackage{amsmath} 
\usepackage{amssymb}  

\usepackage{graphicx}
\usepackage{float}
\usepackage{capt-of}
\usepackage{etoolbox}
\usepackage{adjustbox}
\usepackage{hyperref}

\title{\LARGE \bf 
The Earth ain't Flat: Monocular Reconstruction of Vehicles on Steep and Graded Roads from a Moving Camera
}

\author{Junaid Ahmed Ansari$^{1}$$^{*}$, Sarthak Sharma$^{1}$$^{*}$, Anshuman Majumdar$^{1}$$^{}$, J. Krishna Murthy$^{2}$ and K. Madhava Krishna$^{1}$
\thanks{*The first two authors contributed equally to this work.}
\thanks{$^1$Junaid Ahmed Ansari, Sarthak Sharma, Anshuman Majumdar, and K. Madhava Krishna are with the Robotics Research Center, KCIS, IIIT Hyderabad, India. {\tt\small junaid.ansari@research.iiit.ac.in}}
\thanks{$^2$J. Krishna Murthy is with Montreal Institute of Learning Algorithms (MILA), Universite de Montreal, Canada.}
}

\begin{document}

\makeatletter
\let\@oldmaketitle\@maketitle
\renewcommand{\@maketitle}{\@oldmaketitle
\centering
\includegraphics[width=14.5cm,height=8cm]{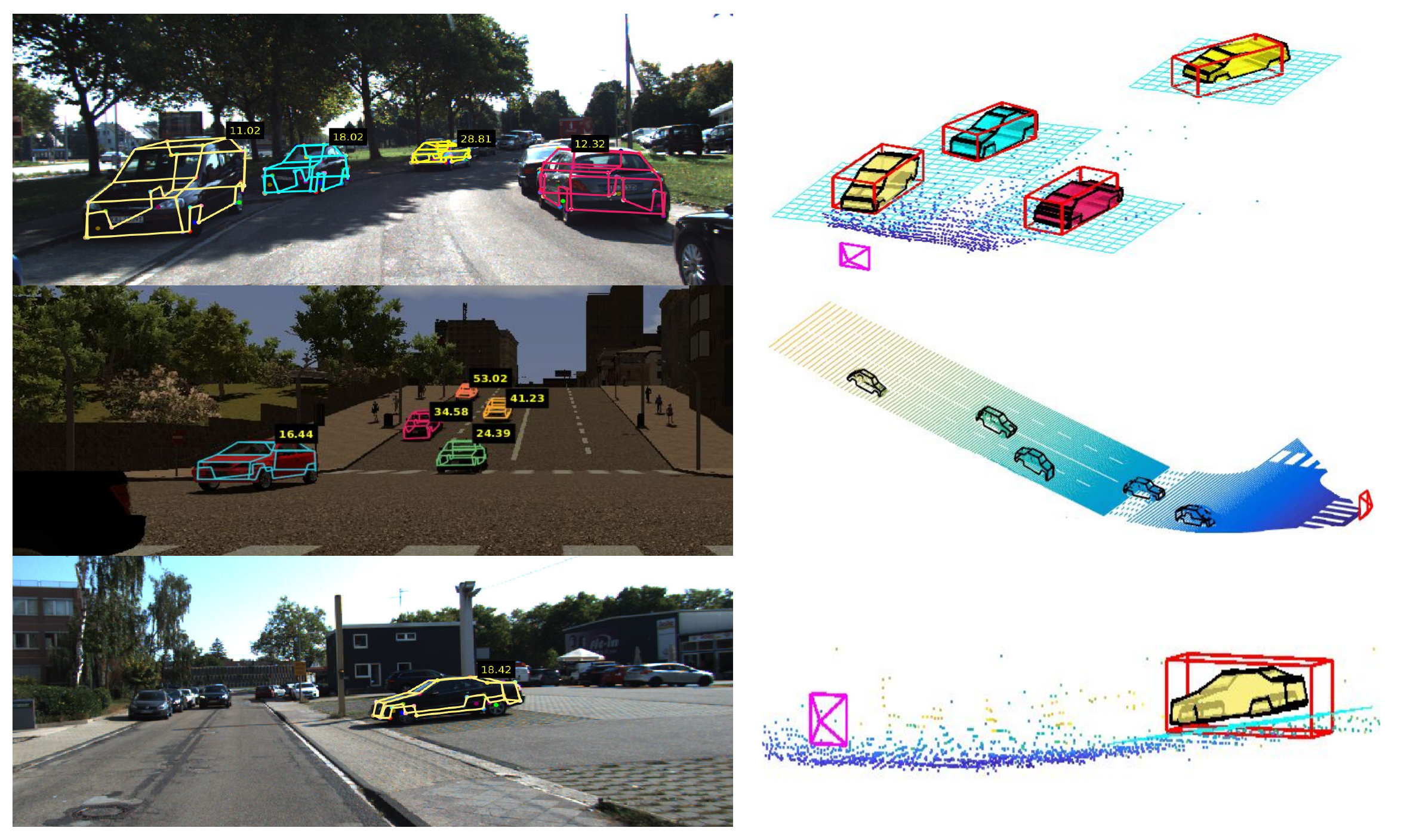}
\vspace{-0.1cm}
\captionof{figure}{Some results showcasing the efficacy of the proposed monocular object localization system. The system is capable of estimating the shape and pose (without scale-factor ambiguity) of objects located on surfaces that do not share the same plane with the moving monocular camera. The images of the scenes contain the projection of the estimated shapes (wireframes) of cars. On the top of each car, we indicate the distance of the car from the camera (in meters). To the right side of each scene, lies the visualization of the estimated wireframe and road points in 3D. For the first and third scenes, we visualize the wireframes with their respective ground truth 3D bounding boxes (shown in red) on the right, highlighting the accurate localization of the objects. In the second scene, we show the accurately estimated cars in 3D, overlayed on a dense ground truth 3D point cloud. Even the objects at over 50 meters distance on steep slopes are accurately localized.}}
\label{fig:teaser}
\makeatother

\maketitle

\vspace{-0.5cm}
\begin{abstract}
Accurate localization of other traffic participants is a vital task in autonomous driving systems. State-of-the-art systems employ a combination of sensing modalities such as RGB cameras and LiDARs for localizing traffic participants, but most such demonstrations have been confined to plain roads. We demonstrate, to the best of our knowledge, the first results for monocular object localization and shape estimation on surfaces that do not share the same plane with the moving monocular camera. We approximate road surfaces by local planar patches and use semantic cues from vehicles in the scene to initialize a local bundle-adjustment like procedure that simultaneously estimates the pose and shape of the vehicles, and the orientation of the local ground plane on which the vehicle stands as well. We evaluate the proposed approach on the KITTI and SYNTHIA-SF benchmarks, for a variety of road plane configurations. The proposed approach significantly improves the state-of-the-art for monocular object localization on arbitrarily-shaped roads.

\end{abstract}


\section{Introduction}
\label{sec:introduction}

With the advent and subsequent commercialization of autonomous driving, there is an increased interest in monocular object localization for urban driving scenarios. While recent monocular localization methods \cite{KM_ICRA,KM_IROS} achieve better localization precision when compared with stereo methods, they are confined to scenarios where the road is (very nearly) flat. This holds true for other monocular object localization systems as well \cite{chandraker2015,mobileye}.

Reconstruction of vehicles from a monocular camera is a challenging task, owing to several factors viz. dearth of stable feature tracks on moving vehicles, self-occlusions, and it is ill-posed if the camera itself is in motion. To overcome some of these, discriminative features \cite{hourglass} and shape priors \cite{KM_IROS,Zia} have been used to pose a bundle adjustment like scheme \cite{KM_IROS} that solves for shape and pose of a detected vehicle, assuming a prior on the shapes of all instances from a category. Using shape priors results in a richer representation of reconstructed vehicles; they are now reconstructed as 3D wireframes rather than 3D bounding boxes.

We present, to the best of our knowledge, the first results for monocular object shape and pose estimation on surfaces that do not share the same plane with the moving monocular camera. We approximate road surfaces by local planar patches and use semantic cues from vehicles in the scene to initialize a local bundle-adjustment like procedure that simultaneously estimates the pose and shape of the vehicles, and the orientation of the local ground plane on which the vehicle stands as well. Using the proposed approach, we accurately reconstruct vehicles, predominantly using cues from only a single image. The presented method works across a variety of road geometries and demonstrate substantial improvements in terms of vehicle localization accuracy on extremely steep and non-planar roads.

To evaluate our approach, we use the popular KITTI \cite{KITTI} and SYNTHIA-SF \cite{SYNTHIA} benchmarks. While sequences from the KITTI \cite{KITTI} dataset only have mild-to-moderate slopes and banks, it provides a fair comparison with other baseline methods \cite{KM_ICRA,KM_IROS}. SYNTHIA-SF \cite{SYNTHIA}, on the other hand, has extremely steep roads and demonstrates the efficacy of the proposed approach in adapting to a wide range of road surfaces.


The remainder of the paper is organized as follows. Section \ref{sec:relatedwork} briefly discusses relevant work on monocular object localization and reconstruction in urban driving scenarios. The proposed approach is outlined in section \ref{sec:approach}. In section \ref{sec:results}, we present an evaluation of the proposed approach on popular benchmarks and discuss the results obtained thereof. Section \ref{sec:conclusions} concludes the paper.

\begin{figure*}[t]
\centering
\includegraphics[width=0.9\textwidth]{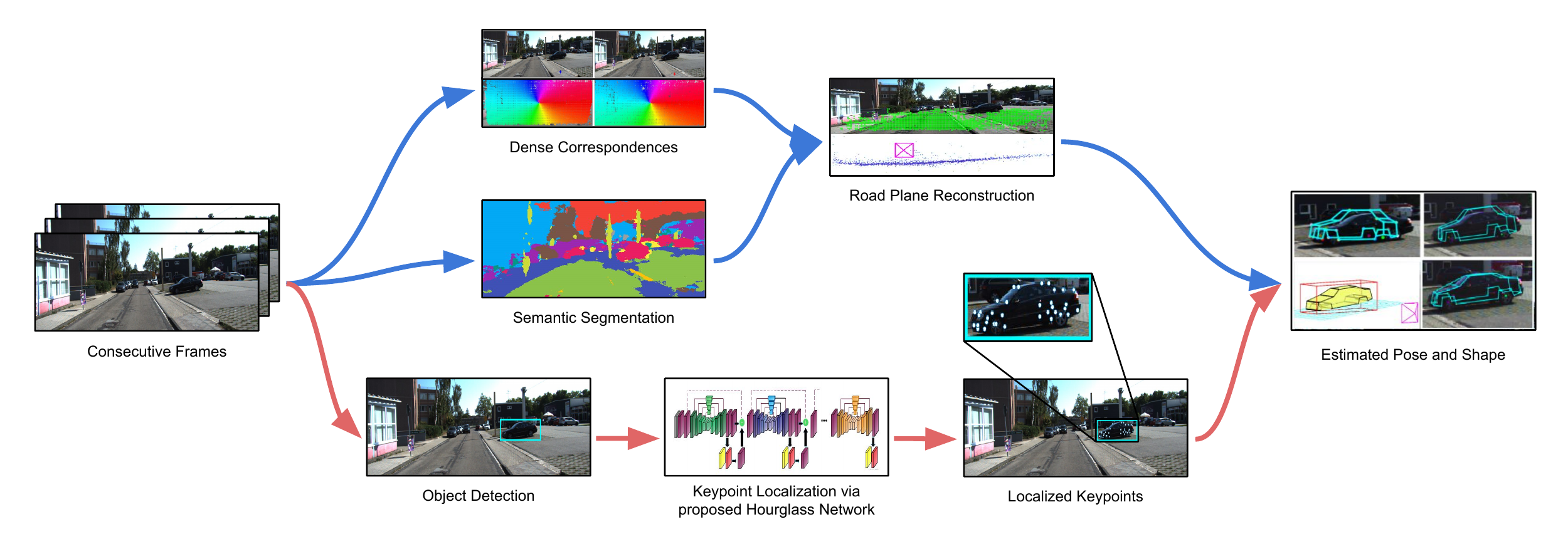}
\caption{Illustration of the proposed pipeline. The system takes as an input, 3 consecutive frames (in case of no lane markers). In the upper half (blue arrows), we illustrate the method for estimating the ground plane i.e. using dense correspondences over the frames and then performing bundle adjustment. In the lower half (red arrows), the detected bounding boxes in each frame are processed using the proposed keypoint localization CNN to obtain 2D locations of a discriminative set of semantic parts. The pose and shape of the object are then adjusted by incorporating the estimated ground plane information.}
\label{fig:pipeline}
\end{figure*}

\section{Related Work}
\label{sec:relatedwork}

In this section, we briefly review relevant literature and contrast it with the proposed approach. 

\subsection{Shape Priors}

Shape priors have been widely used in \cite{Zia,tulsiani_PAMI,KM_IROS} to ease the task of object reconstruction. The underlying hypothesis is that the shape of any instance from a category can be represented as a linear combination of deformations of the mean shape for the category along certain directions, referred to as \emph{basis vectors}. This linear subspace model was used to formulate a stochastic hill climbing problem in \cite{Zia} to estimate the shape and pose of a vehicle in a single image. However, this is prohibitively slow to be used in real-time. 

\subsection{Monocular Localization in Urban Driving Scenarios}

Estimating the 3D shape and pose from a single image has attracted a lot of interest in recent years, supported with the availability of datasets such as KITTI \cite{KITTI}, ShapeNet \cite{ShapeNet} etc. 

Approaches such as \cite{KM_ICRA}, \cite{zhou2016sparseness} follow a 3D-2D pipeline that involves modeling the 3D shape offline and then solving for the 3D deformations in that shape using localized 2D keypoints in RGB image as evidence, thus overcoming the need to explicitly estimate the 3D keypoints. In \cite{KM_ICRA}, an approach to estimate the 3D shape and pose of the vehicles from a single image is presented. The 3D shape of an instance was modeled using a shape prior based on a linear subspace model and deformation coefficients were estimated by solving an optimization problem using vehicle keypoints localized in 2D using a CNN.

In \cite{chandraker2015}, \cite{song2014robust} the authors develop a real-time monocular SfM system leveraging information from multiple image frames. However vehicles are represented as 3D bounding boxes. It was demonstrated in \cite{KM_IROS} that having a richer representation for the vehicle, such as a 3D wireframe, significantly boosts localization accuracy. Mono3D \cite{mono3d} trains a CNN that jointly performs object detection in 2D and in 3D space and estimates oriented bounding boxes for vehicles. Although it outperforms stereo competitors, it made the assumption of a planar road surface.

Similarly, \cite{KM_ICRA,chandraker2015,song2014robust} rely on the assumption that the plane of the vehicle to be localized is coplanar with the plane of the ego car. Most of these methods use the approach outlined in \cite{mobileye} to estimate the depth to a vehicle under the co-planarity assumption.

\subsection{Monocular Road Surface Reconstruction}

There is relatively little work on road surface estimation from a monocular camera. In \cite{roadedge}, the authors propose a simple road edge prediction framework using edges and lanes detected in earlier frames. No surface level reconstruction is provided. In \cite{roadterrain}, road width and shape of the drivable area are estimated using a Conditional Random Field (CRF).

In contrast to the above approaches, the proposed approach is independent of the road plane profile and accurately localizes the vehicle independent of its coplanarity with the ego vehicle. The cost functions provided are robust, fast, and easy to implement; resulting in very accurate shape and pose estimation of the vehicle independent of the plane on which the vehicle is located. The method outperforms the current best competitor \cite{KM_IROS} by a significant margin, highlighting how the existing approaches fail when presented with non-planar road surfaces.

\section{Geometry and Object Shape Costs}
\label{sec:approach}

In this section, we outline our approach to reconstruct vehicles on arbitrarily oriented roads surfaces.

\subsection{Background: Shape Priors}

Along the lines of \cite{Zia,KM_ICRA,KM_IROS}, we assume that each vehicle (in this case, a car) is represented in 3D by a wireframe consisting of $K$ vertices (we use $K = 36$, according to the setup in \cite{Zia}), each of which has a unique semantic meaning. For instance, these vertices could be locations of headlights, tail lights, wheel centers, rooftop corners, etc. that are easily identifiable across all cars. We use a set of aligned 900 CAD models of cars from the ShapeNet \cite{ShapeNet} repository and annotate each of them with $K$ keypoint locations in 3D. We then use the render pipeline presented in \cite{Parv_ICRA} to synthesize a dataset comprising about 2.4 million images of rendered cars with annotated 2D keypoint locations. Over this dataset, we train a keypoint localization network based on the stacked-hourglass architecture \cite{hourglass}. We use this CNN, trained entirely on synthetic data, across all experiments reported in this work. We observe that the network generalizes well to real data, consistent with the findings in \cite{RenderForCNN}.

Using notation from \cite{KM_IROS}, we denote the mean wireframe for the vehicle category by $\bar{X} \in \mathbb{R}^{3K}$. The basis vectors are stacked into a $3K \times B$ matrix denoted $V$. The deformation coefficients (also referred to as the shape parameters) $\Lambda \in \mathbb{R}^{B}$ uniquely determine the shape of a particular instance. If we assume that the object coordinate frame has a rotation $R \in SO(3)$ and translation $t \in \mathbb{R}^3$ with respect to the camera center, any instance $X$ can then be parameterized by the shape prior model as 

\begin{equation}
\centering
X = \hat{R} \left( \bar{X} + V \Lambda \right) + \hat{t} 
\label{eqn:shapePrior}
\end{equation}

Here, $\hat{R} = diag([R, R, ..., R]) \in \mathbb{R}^{3K \times 3K}$, and $\hat{t} = \left(t^T, t^T, ..., t^T\right)^T \in \mathbb{R}^{3K}$. $\bar{X} = \left(\bar{X}_1^T, \bar{X}_2^T, ..., \bar{X}_K^T\right)^T$ is an ordered collection of the 3D locations of the keypoints in the mean wireframe.

If we denote the locations of an ordered collection of 2D keypoints by $\hat{x} = \left(\hat{x}_1^T, \hat{x}_2^T, ..., \hat{x}_K^T\right)^T \in \mathbb{R}^{2K}$, the pose ($R, t$) and shape ($\Lambda$) of the vehicle can be obtained by minimizing the following objective function in an alternating fashion - once for pose, and once for shape.

\begin{equation}
\underset{R, t, \Lambda}{\text{min}} \mathcal{L}_{r} = \| \pi_K \left( \hat{R} \left( \bar{X} + V \Lambda \right) + \hat{t} ; f_x, f_y, c_x, c_y \right) - \hat{x} \|_2^2
\label{eqn:poseAndShapeAdjustment}
\end{equation}

$\pi_K()$ is a vectorized version of the perspective projection operator, which takes in $K$ 3D points and computes their image coordinates, given the camera intrinsics $\mu = (f_x, f_y, c_x, c_y)$. Specifically, $\pi_K$ is the following function.

\begin{equation}
\centering
\begin{aligned}
\pi \left( (X,Y,Z)^T ; \mu \right) = \left( \begin{matrix}
{f_x X \over Z} + c_x \vspace{0.25cm} \\ {f_y Y \over Z} + c_y
\end{matrix} \right) \\
\pi_K\left( (X_1^T, ..., X_K^T)^T ; \mu \right) = \left( \pi(X_1;\mu)^T, ..., \pi(X_K;\mu)^T \right)^T
\label{eqn:projection}
\end{aligned}
\end{equation}

\subsection{System Setup}

We operate on image streams captured by a front-facing monocular (RGB) camera mounted on a car. The height $H$ above the ground at which the camera is assumed to be known a priori (this helps in resolving scale-factor ambiguity in monocular reconstruction). 

We assume that, on each incoming image, an object detector \cite{RRC} runs and detects vehicles in the image (as bounding boxes). We also perform a semantic segmentation of the input image using the SegNet \cite{segnet} convolutional architecture. The proposed pipeline is illustrated in Fig\ref{fig:pipeline}

\subsection{Reconstruction of Vehicles on Slopes}

To formulate a lightweight, yet robust optimization problem for reconstructing vehicles on non-planar road surfaces(i.e. roads with slopes and banks), we assume that the road is locally planar. By this, we mean that the patch of the road that lies exactly beneath a detected vehicle is assumed to be a planar patch. This assumption is corroborated by \cite{chandraker2015}, where allowing each vehicle to have an adaptive local ground plane boosts localization accuracy. 

Each detected vehicle $v$ is on a planar patch parameterized by $({n^v_g}^T,d^v_g)$, where $n^v_g$ is a vector that denotes the normal to the planar patch and $d^v_g$ denotes the distance of the planar patch from the origin of the world coordinate frame.

\subsection*{Resolution of Scale-Factor Ambiguity}

Monocular camera setups inherently suffer from scale-factor ambiguity, i.e., any 3D length estimated from a set of images is accurate up to a positive scalar. But, for the autonomous driving applications, we require that vehicles are localized in \emph{metric scale}, i.e., in real-world units (such as meters, for instance). We resolve scale ambiguity using one of the following two approaches.

\subsubsection*{Using Dimensions of Detected Lanes}

Most roads have lane marking or zebra crossings of standard dimensions that are known to us a priori. We use the method from \cite{lanedetection} to detect lane markings, and if we know the height of the camera above the ground and the dimensions of the lane markings, we can retrieve the planar patch comprising the lane marking and the distance to that lane marking (in meters). Such a method estimates the local ground plane (of a lane marking near the vehicle) using information from just a single image.

\subsubsection*{Using 3-View Reconstruction and Camera Height}

The above method can only be employed on roads where there are lane markings and in particular only if a lane marking is detected near a vehicle, which is not true for all scenarios we encounter. In the more general case, we can recover absolute (metric) scale by using the following 3-view reconstruction scheme. Assume we have three consecutive frames $f_1, f_2, f_3$ with sufficient parallax. We use DeepMatching\cite{deepmatching} for establishing dense correspondences between frames $f_1$ to $f_2$. Then, using a sufficient mix of road and non-road points, we estimate the egomotion between the frames using standard multi-view motion estimation techniques \cite{mvg}. Using the estimated egomotion, we triangulate points \emph{close}\footnote{We expand the car bounding box by a factor of 1.9 to 2.0, and pick all points from the expanded bounding box that are classified as \emph{road} by SegNet \cite{segnet}.} to the car that lie on the road surface and add points from frame $f_3$ to the reconstruction\footnote{This is typically done by propagating feature matches from frame $f_2$ to frame $f_3$, and running a resection routine to estimate the egomotion between frame $f_1$ and frame $f_3$, and then triangulating points from $f_3$ onto the initial reconstruction \cite{mvg}}. A local ground plane patch can then be estimated by estimating a dominant plane from the obtained point cloud using a RANSAC-like routine. Once such a plane is obtained, we can scale the reconstruction such that the median of the Y-coordinates of the estimated plane is roughly equal to the height of the camera above the ground (which is assumed to be known during initial setup).

\begin{figure}[t]
\centering
\includegraphics[width=0.8\linewidth]{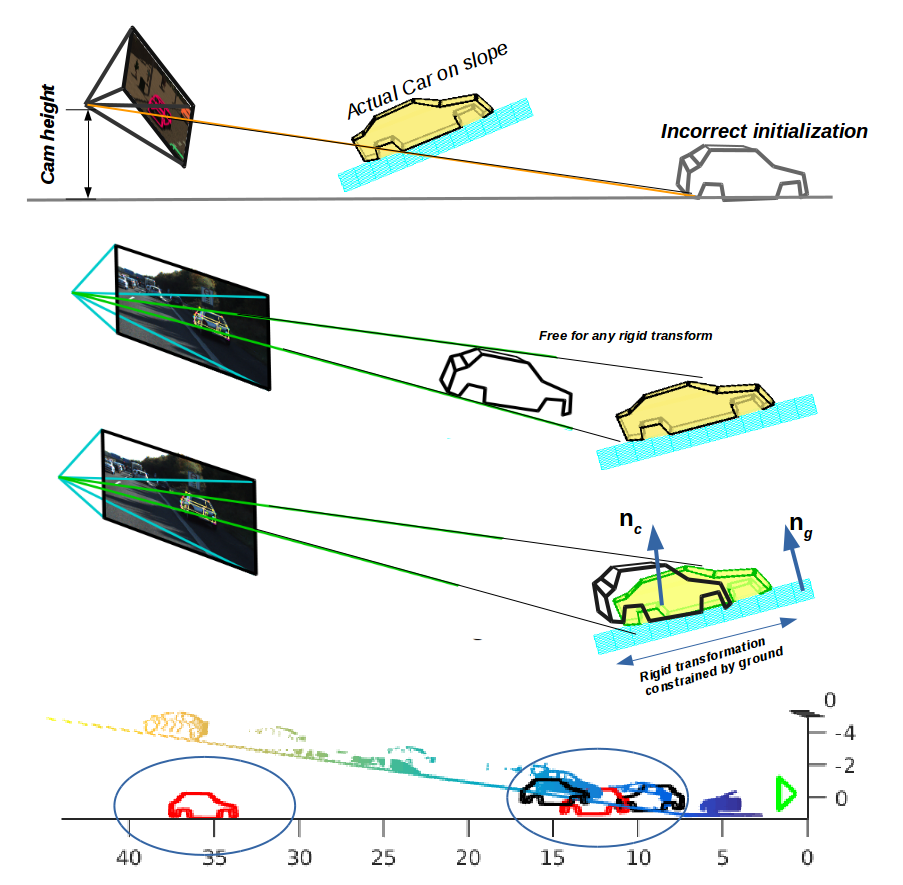}
\caption{From top to bottom - (i) Illustrating how co-planarity assumption results in incorrect initialization in existing approaches (ii) Relying only on minimizing the reprojection error, leaves the optimizer free to rigidly transform the mean car (iii) Joint optimization constrains the car to be on ground while minimizing the reprojection error, resulting in more accurate reconstruction and localization ($n_c$ and $n_g$ are car base and road plane normals respectively) (iv) Failure of co-planarity assumption for steep roads on SYNTHIA-SF \cite{SYNTHIA}. Notice the incorrect initialization of the car on slopes via method of \cite{KM_ICRA} shown in red. Our method is not bound by this co-planarity assumption and initializes the vehicle correctly, shown in black. We overlay the initialized wireframe on the ground truth 3D points for comparison.}
\label{fig:whyGroundHelps}
\end{figure}

\subsubsection*{How does the Ground Plane help?} 

In scenarios where the plane of the vehicle is not same as the plane of the ego car, current methods of estimating shape and pose of the vehicle suffer due to their co-planarity assumption. We circumvent this failure by estimating the ground plane on which the vehicle is located as proposed in Fig. \ref{fig:whyGroundHelps}. The estimation of the ground plane parameters not only helps in correct initialization of the car, but also helps in correct localization of the vehicle constraining it to move on the plane, rather than in the line of sight of the camera to minimize the re-projection error.

\subsection*{Joint Optimization for Ground Plane and Vehicle Pose and Shape Estimation}

Equation \ref{eqn:poseAndShapeAdjustment} represents the optimization problem that is solved to estimate the shape and pose of a vehicle from just a single image or from a pair of images whenever available \cite{KM_IROS}. However, this formulation assumes co-planarity of the ego car and of the object being reconstructed. We illustrate in Fig. \ref{fig:whyGroundHelps} that drastic errors in localization result when the assumption does not hold. 

We assume that, in the current frame, a set of vehicles $\mathcal{V}$ have been detected by the object detection network \cite{RRC}. For a particular vehicle $v \in \mathcal{V}$, we let $X^v_i$ denote the coordinates of the $i^{th}$ keypoint of the vehicle in 3D. Also, we parametrize the local ground plane beneath $v$ by its normal vector $n^v_g$ and the distance of the plane from the origin $d^v_g$. Also, we denote by $n^v_c$ the normal of the car. The normal of the car is defined as the normal of a plane that \emph{best}\footnote{Although, in practice, all 4 wheel centers of a car are co-planar, it may still be numerically hard to determine a plane equation that satisfies all 4 points. So, we fit a plane in the least squares sense to the 4 wheel centers.} fits the keypoints corresponding to the wheel centers of the cars.

We now formulate a set of cost functions that relax the co-planarity assumptions in \cite{KM_ICRA,KM_IROS} and estimate the vehicle's pose and shape as well as the equation of the ground plane patch beneath it.

\subsubsection*{\textbf{Ground Plane Estimation}} We define a ground plane estimation loss term, which \emph{encourages} the vehicle to be as close to the ground plane as possible. Specifically, we obtain the translation vector $t^v_c$ to the bottom of the vehicle $v$\footnote{We first obtain the rigid-body transform to the origin of the vehicle coordinate frame, and then concatenate to it the rigid-body transformation from the origin of the vehicle coordinate frame to the bottom of the vehicle.} from the world origin (typically the camera center). This obtained quantity, in the ideal setting, represents the position vector of a point on the ground plane, the points of which are denoted as $X^v_g$. Formally, this term (for all vehicles in the image) can be represented as follows.

\begin{equation}
\centering
\mathcal{L}_{g} = \sum_{v \in \mathcal{V}} \| n^v_c \cdot t^v_c - d^v_g \|^2
\label{eqn:loss_ground}
\end{equation}

\subsubsection*{\textbf{Normal Alignment}} The normal alignment loss term stipulates that the normal of the vehicle $(n^v_c)$ must be encouraged to be parallel to the normal of the estimated ground plane. An initial guess for the ground plane normal is obtained as described earlier, using either lane markings, or a 3-view reconstruction. This loss can be denoted as follows. $\times(.,.)$ denotes the vector cross product.

\begin{equation}
\centering
\mathcal{L}_{n} = \sum_{v \in \mathcal{V}} \| \times (n^v_c, n^v_g) \|^2
\label{eqn:loss_normal}
\end{equation}

\subsubsection*{\textbf{Disambiguation Prior}} The above loss term has one drawback in that, it is minimized even when the estimated ground plane and vehicle normals are anti-parallel. To disambiguate such unwarranted solutions, we make use of the fact that even the steepest roads in the world have slopes less than $25 \deg$ \cite{roadgrade}. Whenever multiple solutions are avaliable, we encourage the solution that's \emph{more upright} to have a lower cost. If $e_2$ denotes the Y-axis of the camera coordinate system (i.e., the axis vertically pointing down), we formulate the disambiguation prior as follows ($\epsilon$ is a tiny positive constant that provides numerical stability).

\begin{equation}
\centering
\mathcal{L}_{d} = \sum_{v \in \mathcal{V}}  \left\lvert\left\lvert {- 1 \over e_2 \cdot n^v_c + \epsilon } \right\rvert\right\rvert^2 + \left\lvert\left\lvert {- 1 \over e_2 \cdot n^v_g + \epsilon} \right\rvert\right\rvert^2
\label{eqn:loss_disambiguation}
\end{equation}

\subsubsection*{\textbf{Base Point Priors}} We also use a loss term that encourages points along the base of the car (this includes keypoints on the car wheel centers, bumpers, etc) to lie as close to the estimated ground plane as possible. If $X_b$ is a keypoint on the car base, and $\mathcal{K}_b$ denotes the set of all keypoints that lie along the base of the car, base point priors are imposed using the following expression.

\begin{equation}
\centering
\mathcal{L}_{b} = \sum_{v \in \mathcal{V}} \sum_{X_b \in \mathcal{K}_b} \| n^v_c \cdot t^v_c - n^v_c \cdot X_b \|^2
\label{eqn:loss_base}
\end{equation}

\subsubsection*{\textbf{Global Consistency}} Although we assume that each vehicle has its own planar ground patch, it is safe to assume that road planes are not susceptible to abrupt change. This is encoded into the global consistency loss term, that encourages the planar ground patch of a vehicle to be consistent with that of other vehicles around it. If $\mathcal{V}^n$ denotes the set of all vehicles within a distance $d$ around vehicle $v$ ($v$ is usually chosen to be $5-7$ meters), the global consistency loss term is as follows.

\begin{equation}
\centering
\mathcal{L}_{c} = \sum_{v \in \mathcal{V}} \sum_{v^n \in \mathcal{V}^n} \| n^v_g - n^{v^n}_g \|^2 + \| d^v_g - d^{v^n}_g \|^2
\label{eqn:loss_consistency}
\end{equation}

\subsubsection*{\textbf{Dimension Regularizers}} We also place priors on dimensions of vehicles that we observe, which provides a well-conditioned problem to work with and leads to better convergence rates. We use regularizers similar to ones proposed in \cite{KM_IROS}, and denote the loss term by $\mathcal{L}_{reg}$.

\subsubsection*{\textbf{Overall Optimization Problem}} The overall minimization problem involving all the energy terms can be posed as follows (cf. Eq \ref{eqn:poseAndShapeAdjustment} \ref{eqn:loss_ground} \ref{eqn:loss_normal} \ref{eqn:loss_disambiguation} \ref{eqn:loss_consistency} \ref{eqn:loss_base}).

\begin{equation}
\centering
\begin{aligned}
\underset{R, t, \Lambda, n^v_g, d^v_g, n^v_c}{\text{min}}  & \mathcal{L}_{total} = \eta_r\mathcal{L}_{r} + \eta_g\mathcal{L}_{g} + \eta_n\mathcal{L}_{n} \\ & + \eta_d\mathcal{L}_{d} + \eta_b\mathcal{L}_{b} + \eta_c\mathcal{L}_c \eta_{reg}\mathcal{L}_{reg}
\end{aligned}
\label{eqn:loss_total}
\end{equation}

Here, $\eta_r$, $\eta_g$, $\eta_n$, $\eta_d$, $\eta_b$, $\eta_c$, and $\eta_{reg}$ are weighing factors that control the relative importances of each of the loss terms. In practice, $\eta_r$, $\eta_g$, $\eta_d$, and $\eta_b$ are more dominant compared to the other terms. The actual values of these weighing factors do not really matter as long as the above terms are properly weighted.

The above problem is minimized using Ceres Solver \cite{ceres-solver}, a nonlinear least squares minimization framework, using a Levenberg-Marquardt optimizer with a Jacobi preconditioner. In addition, each term is composed with a Huber loss function, to reduce the effect of outliers on the solution obtained.

\section{Experiments and Results}
\label{sec:results}

We perform a thorough quantitative and qualitative analysis of our approach on challenging sequences from the KITTI Tracking \cite{KITTI} and SYNTHIA-SF \cite{SYNTHIA} benchmarks. These sequences are chosen such that they capture a diverse class of road plane profiles viz. uphill, downhill, combinations of them, and even banked road planes.
We compare the 3D localization error of the proposed method with the current state-of-the-art monocular competitor \cite{KM_IROS}, and demonstrate significant improvements. Through a series of systematic evaluations, we demonstrate that ground plane estimation is vital for accurate localization on roads surfaces with pitch and banks. We also demonstrate that our method is independent of the road plane profile on which vehicles are to be reconstructed. In other words, unlike others (such as \cite{KM_ICRA,chandraker2015,mono3d}) we do not make any assumptions that the ego car and the car to be reconstructed are on the same road plane.

\subsubsection*{\textbf{Dataset}}
We use the KITTI \cite{KITTI} tracking benchmark to evaluate our proposed method. Sequences numbered 1, 3, 7, 8, 9, 10, 11 and 20, which contain a large number of vehicles located on roads with varying plane profiles, were used for evaluating our approach. But, KITTI \cite{KITTI} has only a limited number of steep slopes and banks. So, we also select about 200 vehicles located on challenging plane profiles from sequences numbered 1, 2, 4, 5 and 6 of the SYTHIA-SF \cite{SYNTHIA} dataset. We evaluate the previous best monocular competitor \cite{KM_IROS} is also evaluated on the same sequences, to ensure fair comparison.

\subsubsection*{\textbf{Keypoint Network Training}}
The proposed network was trained on the Torch framework \cite{torch}, with data comprising about 2.4 million images, generated synthetically using the modified render pipeline presented in \cite{Parv_ICRA}. For training and validation respectively, the generated data was split in a 75-25 ratio. The keypoint network was trained for 7 epochs on NVIDIA GTX TITAN X GPUs, spanning over about 36 hours.


\begin{table*}[!t]
\centering
\caption{Mean Localization Error (Standard Deviation in parenthesis) in meters for the vehicles evaluated using our approach on the KITTI \cite{KITTI} Tracking dataset (Here (\textless$x$ $m$) and (\textgreater$x$ $m$) denote the set of all cars within a ground-truth distance of $x$ meters and beyond the depth of $x$ meters respectively)}
\begin{tabular}{|c|c|c|c|c|}
\hline
Approach & Overall ($m$) & \textless$=15m$ & \textless$=30m$ & \textgreater$30m$ \\
\hline
Murthy et. al. \cite{KM_IROS} & $2.61 \, (\pm 2.23)$ & $1.59 \, (\pm 0.96)$ & $2.52 \, (\pm 2.16)$ & $4.30 \, (\pm 2.83)$ \\
Ours (with co-planarity assumption) & $1.00 \, (\pm 0.77)$ & $0.67 \, (\pm 0.50)$ & $0.94 \, (\pm 0.69)$ & $2.19 \, (\pm 1.18)$ \\
\textbf{Ours (joint optimization)} & $\textbf{0.86} \, (\pm \textbf{0.87})$ & $\textbf{0.55} \, (\pm \textbf{0.50})$ & $\textbf{0.79} \, (\pm \textbf{0.79})$ & $\textbf{2.16} \, (\pm \textbf{1.18})$ \\
\hline
\end{tabular}
\label{table:overall_KITTI}
\vspace{-0.2cm}
\end{table*}

\begin{table*}[!t]
\centering
\caption{Mean Localization Error (Standard Deviation in parenthesis) in meters for the vehicles with challenging road profiles evaluated using our approach on the KITTI \cite{KITTI} Tracking dataset} 
\begin{tabular}{|c|c|c|c|}
\hline
Approach & Overall ($m$) & \textless$=15m$ & \textgreater$15m$ \\
\hline
Murthy et. al. \cite{KM_IROS} & $2.55 \, (\pm 3.16)$ & $2.32 \, (\pm 2.21)$ & $2.92 \, (\pm 3.38)$ \\
Ours (with co-planarity assumption) & $0.95 \, (\pm 0.89)$ & $0.92 \, (\pm 0.68)$ & $1.00 \, (\pm 0.96)$ \\
\textbf{Ours (joint optimization)} & $\textbf{0.67} \, (\pm \textbf{0.66})$ & $\textbf{0.64} \, (\pm \textbf{0.60})$ & $\textbf{0.72} \, (\pm \textbf{0.71})$ \\
\hline
\end{tabular}
\label{table:slope_KITTI}
\end{table*}

\begin{table*}[!t]
\centering
\caption{Mean Localization Error (Standard Deviation in parenthesis) in meters for the vehicles (including challenging road profile) evaluated using our approach on the SYNTHIA-SF \cite{SYNTHIA} dataset} 
\begin{tabular}{|c|c|c|c|c|}
\hline
Approach & Overall ($m$) & \textless$=15m$ & \textless$=30m$ & \textgreater$30m$ \\
\hline
Murthy et. al. \cite{KM_IROS} & $76.34 \, (\pm 94.03)$ & $54.21 \, (\pm 47.93)$ & $66.28 \, (\pm 88.74)$  & $86.40 \, (\pm 99.32)$ \\
Ours (with co-planarity assumption) & $32.03 \, (\pm 45.60)$ & $6.3 \, (\pm 19.17)$ & $21.76 \, (\pm 65.76)$ & $42.31 \, (\pm 25.42)$ \\
\textbf{Ours (joint optimization)} & $\textbf{0.92} \, (\pm \textbf{0.93})$ & $\textbf{0.66} \, (\pm \textbf{0.49})$ & $\textbf{0.82} \, (\pm \textbf{0.76})$ & $\textbf{1.23} \, (\pm \textbf{1.11})$ \\
\hline
\end{tabular}
\label{table:overall_SYNTHIA}
\end{table*}

\subsection{Localization Accuracy}
\begin{figure}[t]
\centering
\includegraphics[width=1.0\linewidth]{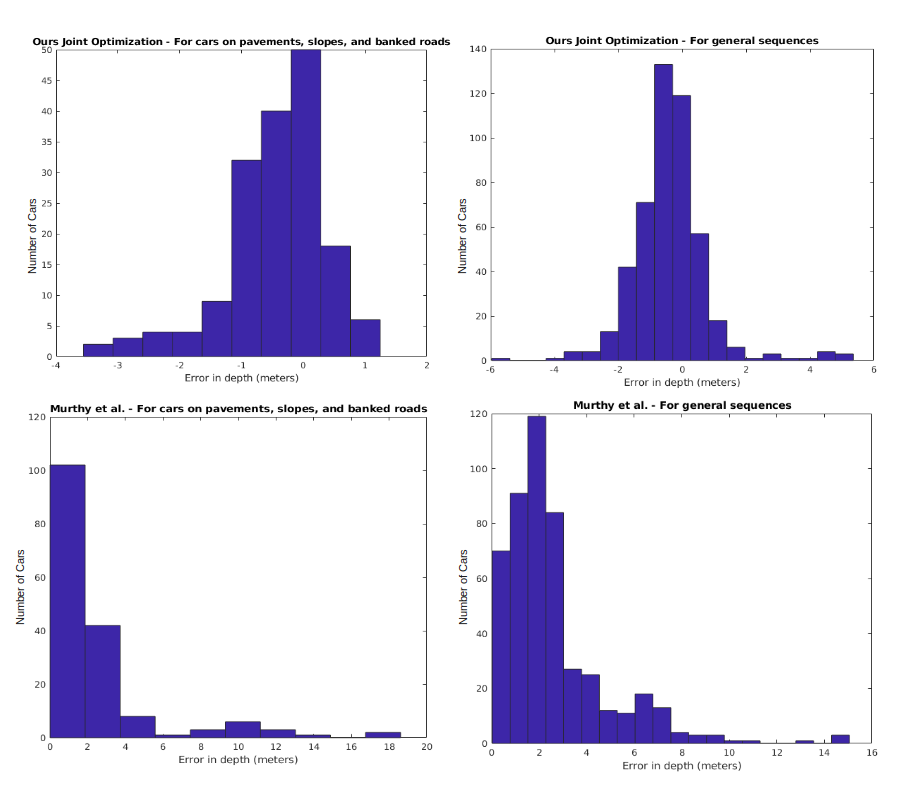}
\caption{Histogram showing the distribution of localization errors. Top (left to right): Plot for sequences from KITTI \cite{KITTI} that exhibit road slant, pitch, and banking. Plot for all evaluated sequences from the KITTI \cite{KITTI} benchmark. These plots show the performance of the proposed approach. The bottom plots are identical, but show the performance of the approach proposed in Murthy et. al. \protect{\cite{KM_IROS}} }
\label{fig:errorHist}
\vspace{-0.5cm}
\end{figure}

\begin{figure}[t]
\centering
\includegraphics[width=0.8\linewidth]{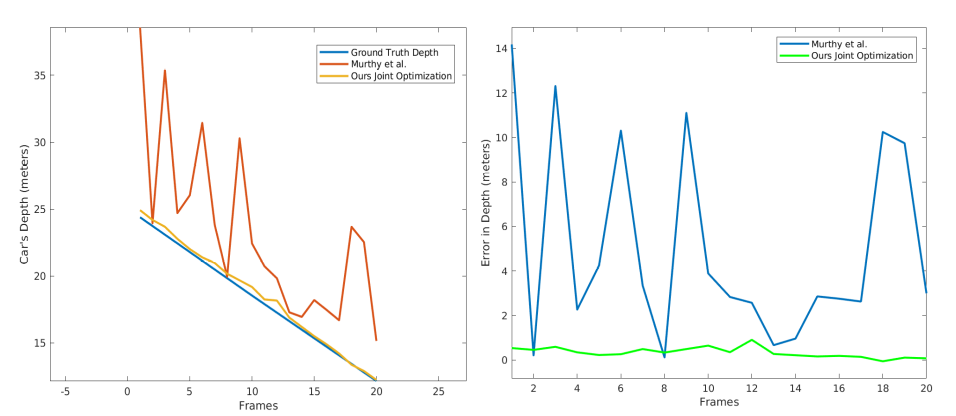}
\caption{\textit{Left}: Predicted depth of a car on a steep slope. We compare predictions with our method with those from \protect{\cite{KM_IROS}} against the ground truth. \textit{Right}: Localization error for the same car when using the proposed method and when using \protect{\cite{KM_IROS}}.}
\label{fig:localizationPlot}
\end{figure}

To evaluate localization precision, we compute the mean Absolute Translational Error (ATE) of the vehicles (in meters) of the approaches considered against the available ground truth information. We present these results in Table \ref{table:overall_KITTI}, Table \ref{table:slope_KITTI} and Table \ref{table:overall_SYNTHIA}. While Table \ref{table:overall_KITTI} captures the overall performance of our approach on KITTI \cite{KITTI} dataset, Table \ref{table:slope_KITTI} presents an analysis of the performance of our approach on KITTI sequences with cars on roads with some pitch or banking angle, or parked on pavements. In Table \ref{table:overall_SYNTHIA}, we perform a thorough analysis of our approach on SYNTHIA-SF \cite{SYNTHIA} which has extremely steep roads, and demonstrate the efficacy of the proposed approach in adapting to a wide variety of road plane profiles.

We outperform the current best monocular localization result of \cite{KM_IROS} on the KITTI benchmark by a significant margin. It is important to note that in \cite{KM_IROS}, the shape priors comprised $14$ keypoints per vehicle, whereas we use a different shape prior model comprising $36$ keypoints per vehicle. However, to emphasize that this improvement does not stem from more expressive shape prior used in this work, we re-implement the approach in \cite{KM_IROS} using our learnt shape priors and provide an ablation study to further drive the point home. This highlights the importance of the inclusion of ground plane in localization. As shown in Table \ref{table:overall_KITTI}, we achieve a mean localization error of 0.86 meters, as compared to 2.61 meters in \cite{KM_IROS}. This is a mark improvement stemming from the inclusion of ground plane.

\begin{figure*}[!t]
\centering
\includegraphics[width=0.8\textwidth]{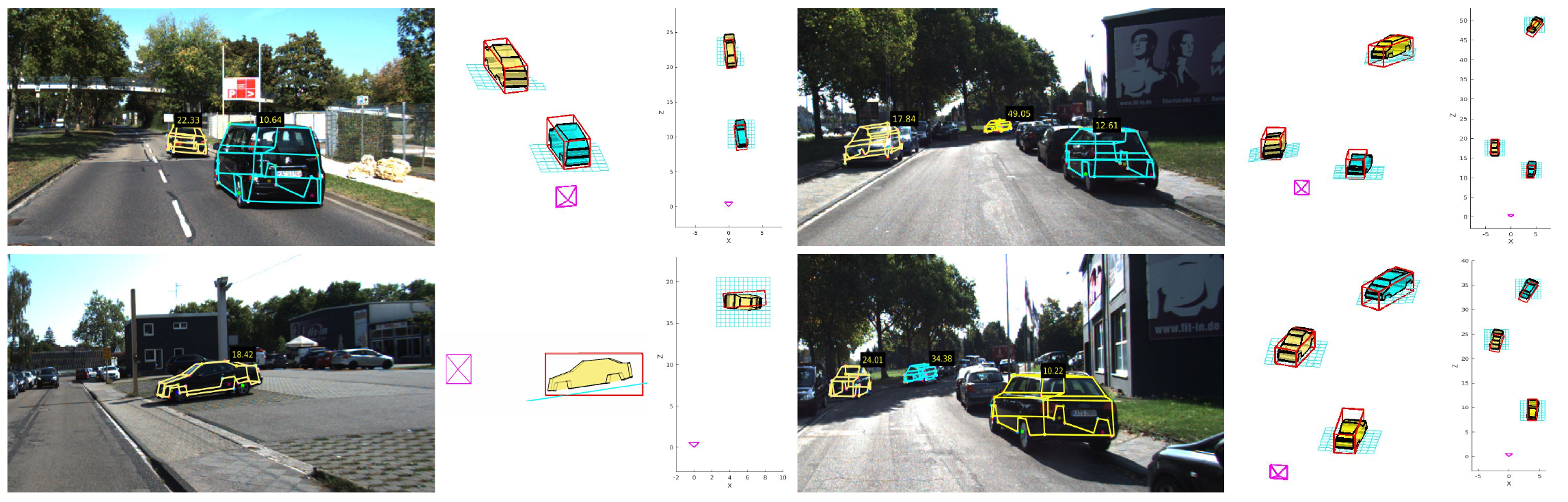}
\caption{Qualitative results on KITTI (with static and moving cars). The images of the scenes contain the projection of the estimated shapes (wireframes) of the cars. On the top of each car, it's depth w.r.t. the camera is displayed. Beside each scene, lies the visualization of the estimated wireframe in 3D, and the bird's eye view of the cars, along with the overlayed ground truth bounding box (in red).}
\label{fig:qualkitti}
\end{figure*}

\begin{figure*}[!t]
\centering
\includegraphics[width=0.9\textwidth]{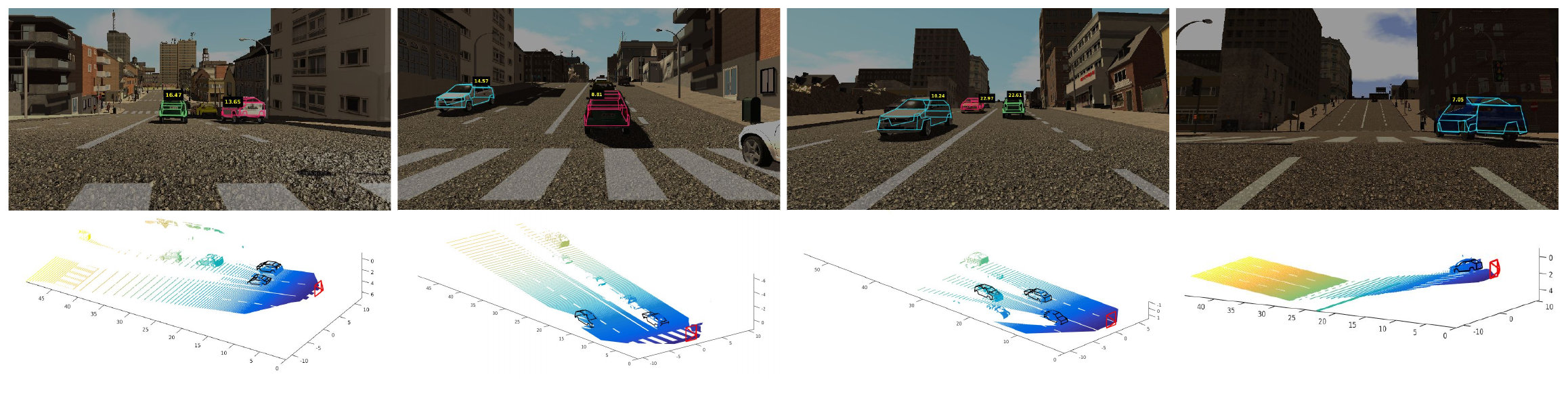}
\caption{Qualitative results on SYNTHIA-SF (with static and moving cars). The images of the scenes contain the projection of estimated shapes (wireframes) of the cars. On the top of each car, depth w.r.t. the camera is displayed. Below each scene image, lies the visualization of the estimated wireframe in 3D, overlayed on a dense 3D point cloud. The proposed method is able to generalize well on cars present on different plane profiles.}
\label{fig:qualsyn}
\end{figure*}

We also address the challenging sequences with moderate slopes on KITTI and provide our localization errors in Table \ref{table:slope_KITTI}, and perform an ablation study of our approach to highlight how our the inclusion of ground plane reduces the localization error to 0.67 meters, as compared to an error of 2.55 meters given by \cite{KM_ICRA}. The current state-of-the-art \cite{KM_ICRA} relies on the assumption that the plane of the vehicle and the ego car is same, i.e they are co-planar. We circumvent this assumption leading to a highly accurate localization of the vehicle irrespective of the fact that it is co-planar with the ego car or not. For vehicles that are close to the car, we achieve a high degree of precision (mean error of about 0.67 meters, with a low standard deviation as well).

To further evaluate our approach, we test it on the extremely challenging SYNTHIA-SF \cite{SYNTHIA} dataset which has steep road surfaces, having various non-planar profiles. \cite{KM_ICRA} fails completely in the task of accurate localization of objects in such scenarios, due to the assumption that the plane of the vehicle and that of the ego car is same, and fails to recover the correct shape and pose. Moreover, the method given by \cite{KM_IROS} fails drastically in non-planar surfaces, giving a mean localization error of 76.34 meters, amplified by the non-generalizable nature of the 14 keypoint network which leads to inaccurate keypoint localizations. Our system achieves a mean localization error of 0.92 meters, the results of which are shown in Table \ref{table:overall_SYNTHIA}. The proposed method generalizes well to different plane profiles and performs significantly well. Once again, we stress the importance of ground plane and exhibit how it's inclusion helps us to perform significantly better as compared to the approach of \cite{KM_ICRA}, which assumes co-planarity of the vehicles and ego car and hence fails in such challenging road profiles. Fig. \ref{fig:errorHist} shows the error distribution of our approach (top two) and for \cite{KM_ICRA} (bottom two). 

\subsection{Keypoint Localization}
To evaluate the accuracy of our 2D keypoint localization network, we use the standard PCK (Percentage of Correct Keypoints) and APK (Average Precision of Keypoints) metrics, used in \cite{VpsKps}, \cite{hourglass} and \cite{RamananPoseEstimation}. A very tight threshold of 2 pixels is used in our experiments and analysis, for the determination of the correctness of our keypoint estimate. Our trained keypoint model achieved a PCK measure of $\textbf{96.89\%}$ at $\alpha=0.1$ APK, on the aforementioned validation set. The network was deployed on KITTI and SYNTHIA-SF datasets.

\subsection{Qualitative Results}
We showcase the qualitative results of our approach on challenging KITTI and SYNTHIA-SF scenes with moderate to high slopes. For KITTI, in Fig. \ref{fig:qualkitti}, we overlay the final estimate of the car in 3D along with the ground truth 3D bounding box to show how our approach estimates the vehicle shape and pose accurately. For SYNTHIA-SF, in Fig. \ref{fig:qualsyn}, we overlay the estimate of the car after shape and pose adjustment on the ground truth scene points to highlight the accurate shape and pose estimation of the car.
\vspace{-0.2cm}

\subsection{Summary of Results}
The cornerstone of this effort was to highlight that the presence of non-planar road profiles leads to an unsuccessful pose estimation of cars in urban scenarios by the current state-of-the-art approach, due to the fact that it relies on the co-planarity of the ego car and the vehicle. Our proposed approach is independent of the plane profile on which the car is located. We improve by a large margin through the inclusion of the ground plane in KITTI sequences, which have moderate slopes. We report these results in Table \ref{table:overall_KITTI} and in Table \ref{table:slope_KITTI}. The importance of the proposed approach is highlighted in Table \ref{table:slope_KITTI}, where we achieve a performance boost of about 4 times in scenes with moderate slopes. For an overall comparison on KITTI, we evaluate our approach on scenes with different planar and non-planar road surfaces and show an improvement of about 3 times. We further present the performance of our approach on SYNTHIA-SF \cite{SYNTHIA} which has extremely steep scenes, resulting in a catastrophic failure of the current state-of-the-art monocular shape and pose estimation \cite{KM_ICRA}. Our performance is significantly improved in such scenes, irrespective of the road profiles, the results of which are reported in Table \ref{table:overall_SYNTHIA}. We also perform an ablation study, reported in Table \ref{table:overall_KITTI}, Table \ref{table:slope_KITTI} and Table \ref{table:overall_SYNTHIA}, to highlight the importance of our ground plane estimation policy, and show that it provides a significant performance boost over just the utilization of a well constrained 36 keypoint system.

\section{Conclusions}
In this work, we presented an approach for accurate 3D localization and shape estimation of vehicles on steep road surfaces. Most current monocular localization systems assume co-planarity of the vehicle to be localized and the ego car, for accurate localization. However, since the assumption does not always hold in the real world, we propose the incorporation of ground plane information (and joint estimation of that information). We show that, this works well in practice, as evident by significant improvements over the state-of-the-art monocular localization methods and thus make a strong case for exploiting ground plane information. Future work could work towards building denser models of roads, and focus on heavy traffic situations - where not much of the road surface is visible.
\label{sec:conclusions}


\small{
\bibliography{references}
\bibliographystyle{IEEEtran}
}

\end{document}